\documentclass[]{bytedance}
\usepackage[toc,page,header]{appendix}

\usepackage{minitoc}
\usepackage{amsfonts}
\usepackage{amssymb}
\usepackage{tabularx}
\usepackage{listings}
\usepackage{xcolor}
\usepackage{cancel}

\usepackage{tabulary,multirow,xspace}
\usepackage{fixmath,mathtools,nicefrac,mmstyle}
\usepackage{subcaption}

\usepackage{wrapfig}
\usepackage{multicol}
\usepackage[most]{tcolorbox}
\usepackage{pifont}

\usepackage{booktabs}
\usepackage{multirow}
\usepackage{makecell}

\definecolor{codegreen}{rgb}{0,0.6,0}
\definecolor{codegray}{rgb}{0.5,0.5,0.5}
\definecolor{codepurple}{rgb}{0.58,0,0.82}
\definecolor{backcolour}{rgb}{0.95,0.95,0.92}
\definecolor{boxblue}{RGB}{57,89,163}
\definecolor{boxbluebg}{RGB}{230,237,250} 

\lstdefinestyle{mystyle}{
    backgroundcolor=\color{backcolour},   
    commentstyle=\color{codegreen},
    keywordstyle=\color{magenta},
    numberstyle=\tiny\color{codegray},
    stringstyle=\color{codepurple},
    basicstyle=\ttfamily\footnotesize,
    breakatwhitespace=false,         
    breaklines=true,                 
    captionpos=b,                    
    keepspaces=true,                 
    numbers=none,                    
    numbersep=5pt,                  
    showspaces=false,                
    showstringspaces=false,
    showtabs=false,                  
    tabsize=2
}
\definecolor{mygray1}{gray}{.95}
\definecolor{mygray2}{gray}{.9}
\definecolor{mygray3}{gray}{.95}
\usepackage{pifont}

\newlength\savewidth
\newcolumntype{x}[1]{>{\centering\arraybackslash}p{#1pt}}

\newcommand{\app}{\raise.17ex\hbox{$\scriptstyle\sim$}}

\usepackage{xcolor}
\usepackage{graphicx}
\usepackage{amssymb}
\usepackage{pifont}
\usepackage{floatrow}
\usepackage{amsmath} 
\usepackage{float}
\usepackage{wrapfig}
\usepackage{multirow}
\usepackage{tcolorbox}
\tcbuselibrary{breakable, skins, raster}
\usepackage{listings}
\usepackage{listings}

\usepackage{algorithm}
\usepackage{algorithmic}
\definecolor{myblue}{RGB}{210, 225, 255}
\definecolor{mytextblue}{RGB}{51, 161, 201}
\definecolor{mypurple}{RGB}{218, 112, 214}

\definecolor{commentgreen}{rgb}{0.1, 0.4, 0.1}
\definecolor{keywordblue}{rgb}{0.1, 0.1, 0.7}
\definecolor{stringred}{rgb}{0.7, 0.1, 0.1}

\lstdefinestyle{mystyle}{
    commentstyle=\color{commentgreen},
    keywordstyle=\color{keywordblue},   
    stringstyle=\color{stringred},
    basicstyle=\ttfamily\scriptsize, 
    breaklines=true,
    keepspaces=true,
    showstringspaces=false,
    frame=none,                     
    language=Python, 
}

\newcommand{\name}{Stream-T1}
\title{\name{}: Test-Time Scaling for Streaming \\ Video Generation}

\author{
Yijing Tu$^1$\quad
Shaojin Wu$^3\dagger$\quad 
Mengqi Huang$^1\dagger$\quad
Wenchuan Wang$^1$\quad 
Yuxin Wang$^2$ \\
Chunxiao Liu$^3$\quad
Zhendong Mao$^{1*}$
}

\affiliation[]{$^1$ University of Science and Technology of China \quad $^2$ FrameX.AI \quad $^3$ Independent Researcher\\ $^*$ Corresponding author \quad $\dagger$ Project Lead}

\abstract{
While Test-Time Scaling (TTS) offers a promising direction to enhance video generation without the surging costs of training, current test-time video generation methods based on diffusion models suffer from exorbitant candidate exploration costs and lack temporal guidance. To address these structural bottlenecks, we propose shifting the focus to streaming video generation. We identify that its chunk-level synthesis and few denoising steps are intrinsically suited for TTS, significantly lowering computational overhead while enabling fine-grained temporal control. Driven by this insight, we introduced \textbf{Stream-T1}, a pioneering comprehensive TTS framework exclusively tailored for streaming video generation. Specifically, Stream-T1 is composed of three units: (1) \textbf{Stream‑Scaled Noise Propagation}, which actively refines the initial latent noise of the generating chunk using historically proven, high-quality previous chunk noise, effectively establishes temporal dependency and utilizing the historical Gaussian prior to guide the current generation; (2) \textbf{Stream‑Scaled Reward Pruning}, which comprehensively evaluates generated candidates to strike an optimal balance between local spatial aesthetics and global temporal coherence by integrating immediate short-term assessments with sliding-window-based long-term evaluations; (3) \textbf{Stream‑Scaled Memory Sinking}, which dynamically routes the context evicted from KV-cache into distinct updating pathways guided by the reward feedback, ensuring that previously generated visual information effectively anchors and guides the subsequent video stream. Evaluated on both 5s and 30s comprehensive video benchmarks, Stream-T1 demonstrates profound superiority, significantly improving temporal consistency, motion smoothness, and frame-level visual quality.
}

\date{\today}
\checkdata[Project Page]{\url{https://stream-t1.github.io/}}
\correspondence{Mengqi Huang at \email{huangmq@ustc.edu.cn}}

\begin{document}
\maketitle

\section{Introduction}
\label{sec:Introduction}
The domain of video synthesis has experienced remarkable advancements in recent years. Among current paradigms, streaming video generation\cite{chen2025skyreels,teng2025magi,yin2025slow,huang2025self,yang2025longlive,lu2025reward,li2026rolling} stands out as a highly promising paradigm for synthesizing exceptionally long videos, elegantly integrating the sequential dependency modeling of autoregressive architectures with the high-fidelity visual generation of diffusion models. Typically, the extraordinary capabilities of these streaming models are built upon distilling diffusion models, a process that inherently demands massive datasets and vast computational resources.

Despite these substantial advancements, synthesizing videos that consistently maintain strict semantic alignment, coherent motion, and long-term temporal consistency remains an open challenge. Furthermore, the traditional paradigm of scaling up models during the training phase is hitting a ceiling, heavily constrained by exorbitant costs and resource demands. Recently, inspired by successes in Large Language Models, pioneering works\cite{wu2026imagerysearch,oshima2025inference,he2025scaling,zhao2026latsearchlatentrewardguidedsearch} have introduced Test-Time Scaling (TTS)\cite{zhang2025and} to video generation and have empirically proven that dynamically scaling computational budgets during inference phase offers a highly effective and cost-efficient pathway to boost video generation quality. Despite this promising potential, approach like ImagerySearch\cite{wu2026imagerysearch} rely on video diffusion models to synthesize the entire video simultaneously. This mechanism forces the search process into a global, high-dimensional space. Coupled with the inherent requirement of multi-step denoising, each candidate demands massive computational resources, severely limiting the overall efficiency of the search process. Furthermore, the simultaneous denoising of all frames fundamentally precludes the ability to inject fine-grained guidance along the temporal axis. Consequently, any localized temporal artifact mandates the rejection of the entire video sequence, rendering dynamic temporal correction impossible.

To address the limitations of existing video TTS methods, we shift the focus to Streaming Video Generation. Operating in a chunk-by-chunk autoregressive manner with minimal denoising steps (e.g., 4 steps per chunk), streaming generation is intrinsically aligned with the principles of Test-Time Scaling. In this paper, we introduce \textbf{\textit{Stream-T1}}, a novel Test-Time Scaling framework tailored for streaming video generation. Combined with candidate selection, \textit{Stream-T1} actively optimizes the generation trajectory by dynamically refining both the latent noise and the context memory. First, we design a \textbf{\textit{Stream‑Scaled Noise Propagation}} mechanism that actively refines the initial latent noise of the current chunk using historically proven, high-quality trajectories, anchoring the exploration space to ensure smooth temporal transitions. Second, we formulate a \textbf{\textit{Stream‑Scaled Reward Pruning}} to evaluate generated candidates, establishing a equilibrium between local spatial aesthetics and global temporal coherence. Finally, guided by these precise reward signals, we introduce a \textbf{\textit{Stream‑Scaled Memory Sinking}}. It dynamically routes the context evicted from KV-cache into distinct updating pathways (Discard, EMA-Sink, or Append-Sink) through semantic boundary detection, effectively decoupling short-term continuity from long-term memory preservation, ensuring that previously generated visual information effectively anchors and guides the subsequent video stream, thereby maintaining global consistency and coherence over extremely long horizons.

Extensive experiments on 5s and 30s video generation benchmarks demonstrate that \textit{Stream-T1} establishes new state-of-the-art performance. Compared to strong baselines, our method significantly improves temporal consistency, motion smoothness, and frame-level visual quality.

In summary, our main contributions are threefold:
\begin{itemize}
    \item \textbf{Concept .} We pioneer the exploration of Test-Time Scaling in streaming video generation and propose \textit{Stream-T1}, the first comprehensive framework tailored for this paradigm.By jointly leveraging search algorithms to expand the candidate space and active strategies to refine the generation process, it significantly enhances the overall quality of the generated videos.
    \item \textbf{Technology.} The proposed \textit{Stream-T1} framework consists of three components: (1) \textit{Stream‑Scaled Noise Propagation}, which actively refines the initial latent noise of the generating chunk using historically proven, high-quality previous chunk noise; (2) \textit{Stream‑Scaled Reward Pruning}, which comprehensively evaluates generated candidates to strike an optimal balance between local spatial aesthetics and global temporal coherence; (3) \textit{Stream‑Scaled Memory Sinking}, which dynamically manages the KV-cache updating pathways guided by the reward feedback, effectively preserving long-term semantics and guiding the subsequent video stream. 
    \item \textbf{Performance.} Comprehensive quantitative and qualitative evaluations reveal that \textit{Stream-T1} significantly outperforms existing state-of-the-art baselines, showcasing remarkable long-term stability and visual fidelity in extended video generation.
\end{itemize}

\section{Related Work}
\label{sec:Related Work}

\subsection{Test-time Video generation}
Test-Time Scaling\cite{zhang2025and,snell2025scaling,liu2025can,alomrani2025reasoning,guo2025deepseek,jaech2024openai,muennighoff2025s1,ramesh2025test,he2025scaling,singhal2025general,li2025reflect,zhuo2025reflection,ji2026compositional} boosts the performance of pre-trained models by increasing the computational budget directly during the inference phase. Existing test-time scaling methods primarily operates as search, utilizing feedback mechanisms to select the optimal samples from multiple candidates. For instance, ImagerySearch\cite{wu2026imagerysearch} dynamically adjusts both the inference search space and reward function according to semantic relationships in the prompt. EvoSearch\cite{he2025scaling} reformulates test-time scaling for diffusion and flow models as an evolutionary search problem, leveraging principles from biological evolution to efficiently explore and refine the denoising trajectory. Video-T1\cite{liu2025video} applies TTS to video generation through a frame-by-frame autoregressive paradigm guided by beam search. Nevertheless, high-dimensional video latent search space and the requirement of multi-step denoising dramatically inflate computational costs. Our proposed Stream-T1, tailors TTS for streaming video generation characterized by chunk-level synthesis and few-step denoising. By inherently forming a "shallow search tree with wide branches", our framework maximizes the computational cost-effectiveness.

\subsection{Memory Management in Streaming Video Generation}
As the video duration extends, streaming generation modeling the full conditional probability requires the continuous accumulation of historical information, inevitably leading to severe context overload and computational bottlenecks.To mitigate this memory explosion, existing approaches\cite{huang2025self,yang2025longlive,lu2025reward,li2026rolling} heavily rely on heuristic context management strategies, yet they frequently fall victim to a severe spatial-temporal trade-off. For instance, methods employing naive sliding window attention (e.g., Self-forcing\cite{huang2025self}) aggressively discard early history, which inherently causes global inconsistency and severe quality drift over time. To preserve early context, subsequent works like LongLive\cite{yang2025longlive} incorporate a static attention sink mechanism, however, relying on fixed initial frames fails to capture intermediate semantic changes, often leading to unnatural subject morphing and frame repetition. Even advanced strategies like Reward Forcing\cite{lu2025reward} attempt to compress discarded history via exponential moving average updates, but their indiscriminate fusion of historical states inevitably blurs and corrupts distinct semantic features during sudden motion changes or scene transitions. Our proposed Stream‑Scaled Memory Sinking dynamically routes the context evicted from KV-cache window into distinct updating pathways (Discard, EMA-Sink, or Append-Sink) through semantic boundary detection, effectively decoupling short-term continuity from long-term memory preservation, ensuring that previously generated visual information effectively anchors and guides the subsequent video stream.
\section{Methodology}
\label{sec:Methodology}
\begin{figure*}[h]
  \centering
  \includegraphics[width=\linewidth]{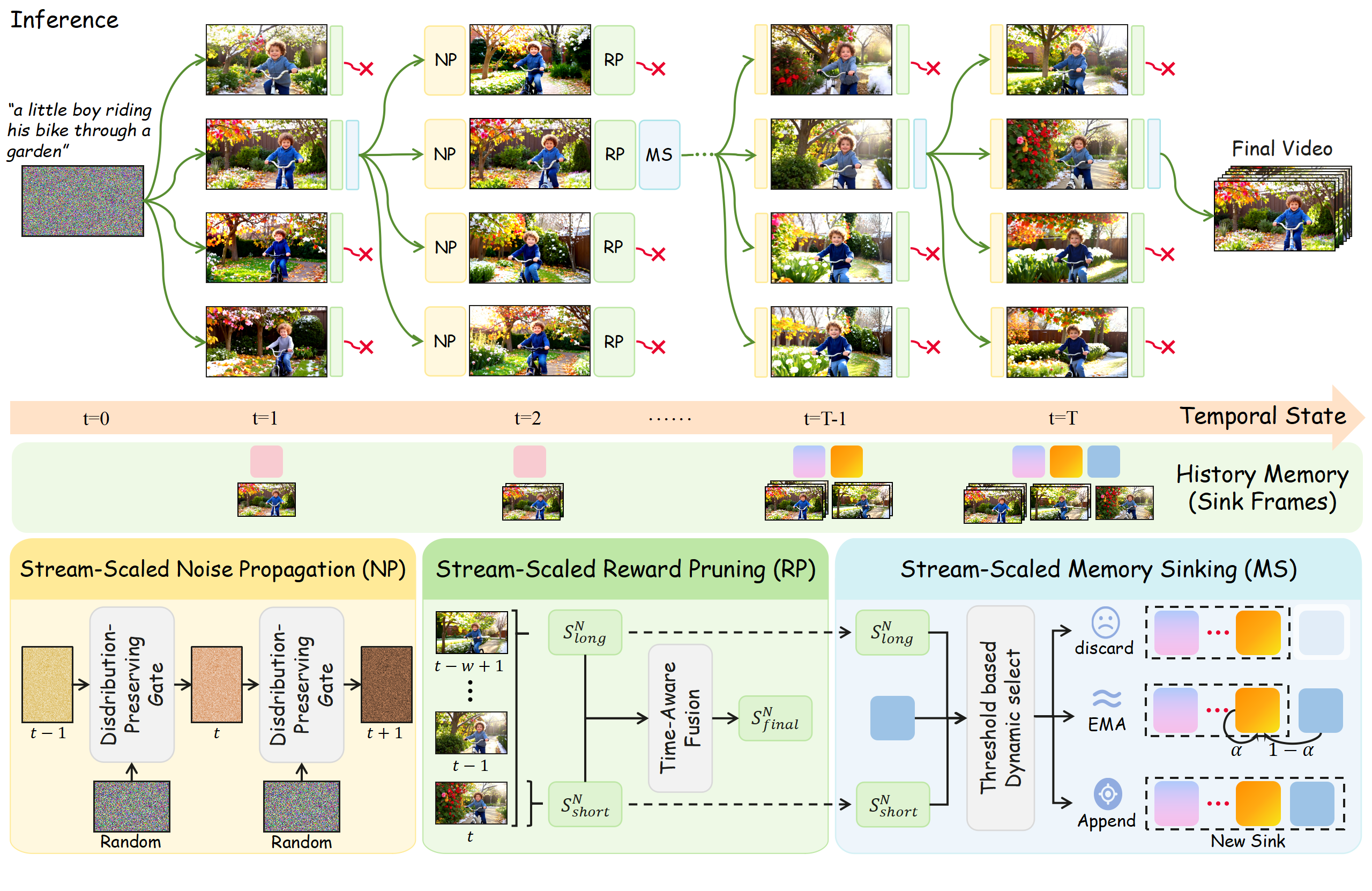}
  \caption{\textbf{Overview of the Stream-T1 chunk-level scaling pipeline. }For each chunk, our test-time scaling mechanism operates in three sequential stages: (1)Stream-Scaled Noise Propagation, where the initial noise latents for the current chunk are actively sampled, conditioned on the optimal noise trajectories retained from the previous chunk. (2)Stream-Scaled Reward Pruning, where fully generated chunk candidates are holistically evaluated via a Long-Short Combined Reward to balance local spatial aesthetics with global temporal coherence. Based on this feedback, suboptimal branches are pruned to propagate only the optimal generation trajectories.(3)Stream‑Scaled Memory Sinking, where the memory sink is adaptively updated based on reward scores to preserve long-term semantics and guide the subsequent video generation. We present the full search trajectory of the illustrated case in the Figure \ref{fig:search}. }
  \label{fig:pipeline}
\end{figure*}

The overall pipeline of our approach is illustrated in Figure \ref{fig:pipeline}. Built upon the LongLive\cite{yang2025longlive}, our framework employs beam search algorithms to systematically expand the candidate space. Specifically, for each autoregressive chunk, Stream-T1 operates through three distinct phases. First, prior to synthesis, \textbf{Stream‑Scaled Noise Propagation} mechanism (Section \ref{sec:noise}) actively refines the initial latent noise of the generating chunk using historically proven, high-quality chunk noise. Second, following the generation, we formulate a \textbf{Stream-Scaled Reward Pruning} (Section \ref{sec:reward}) that comprehensively evaluates generated chunk candidates, establishing an optimal equilibrium between local spatial aesthetics and global temporal coherence. Finally, post-pruning, we introduce an \textbf{Stream‑Scaled Memory Sinking} (Section \ref{sec:memory}), which dynamically routes the context evicted from KV-cache into distinct updating pathways through semantic boundary detection, ensuring that previously generated visual information effectively anchors and guides the subsequent video stream.

Formally, taking the generation of the n-th chunk as a concrete example, we elaborate on our test-time scaling methodology across three sequential stages: pre-synthesis conditional noise initialization, post-synthesis reward-guided pruning, and post-pruning adaptive memory sinking.
\subsection{Preliminary}
\textbf{Autoregressive video diffusion models: }Current text-to-video generation are primarily dominated by diffusion and autoregressive architectures. Video diffusion models\cite{blattmann2023stable,guo2023animatediff,qing2024hierarchical,wang2025wan,singer2022make,wang2023modelscope,yang2024cogvideox,kong2024hunyuanvideo,zhou2022magicvideo,zhang2025show} typically leverage bidirectional attention mechanisms to denoise all frames concurrently. While impressive, this global parallel mechanism inherently restricts the maximum duration of generated videos. Conversely, autoregressive models\cite{kondratyuk2023videopoet,ren2025next,yan2021videogpt,gu2025long,ji2026videoar,deng2024autoregressive,yuanlumos,yu2025videomar} operate on a sequential generation paradigm, predicting tokens based on historical contexts. Recently,  autoregressive Video Diffusion Models\cite{gao2024ca2,li2024arlon,liu2025rolling,weng2024art,yin2025slow,huang2025self} have emerged to elegantly integrate the strengths of both paradigms via temporal autoregressive generation and spatial iterative denoising. Given a text prompt $c$, the chain rule factorizes the joint distribution of the $N$ video frames denoted as $x^{1:N}=(x^1,x^2,...,x^{N})$ into a product of conditional distributions: $p_\theta(x^{1:N} | c) = \prod_{i=1}^N p_\theta(x^i | x^{<i}, c).$ Within this framework, each AR generation step—representing the conditional distribution $p_\theta(x^i | x^{<i}, c)$ is modeled by a few-step denoising diffusion model $G_{\theta}$. Given a defined set of denoising timesteps $\{t_0,t_1,...,t_T\}$, where ${t_T}$ denotes pure noise and ${t_0}$ denotes the clean data, the model generates the $i$-th chunk by progressively denoising an initial Gaussian noise $x^i_{t_T} \sim \mathcal{N}(0, I)$ conditioned on the previously generated history $x^{<i}$. Specifically, at a given timestep $t_j$, the diffusion model $G_\theta$ first predicts the intermediate clean sample $\hat{x}^i_{t_0}$. Subsequently, a forward noising process $\Psi(\cdot)$ is applied to this clean estimate to inject controlled Gaussian noise, yielding the state $x^i_{t_{j-1}}$ for the next step. Thus, the iterative generation process can be formulated as : 
\begin{equation}
p_\theta(x^i | x^{<i}, c) = f_{\theta, t_1} \circ f_{\theta, t_2} \circ \dots \circ f_{\theta, t_T}(x^i_{t_T}),
\end{equation}
where $f_{\theta, t_j}(x^i_{t_j}) = \Psi(G_\theta(x^i_{t_j}, t_j, x^{<i}, c),\epsilon_{t_{j-1}} , t_{j-1})$.

However, as the video duration extends, modeling the full conditional probability $p(x^i | x^{<i})$ leads to severe context overload and computational bottlenecks. Self-forcing\cite{huang2025self} approximates the condition as $p(x^i | x^{i-w+1:i-1})$ with a window size $w$. However, the aggressive discarding of early history inherently causes global inconsistency over time.LongLive models\cite{yang2025longlive} $p(x^i | x^1, x^{i-w+1:i-1})$ by anchoring the initial chunks as a global reference. It heavily relies on static initial context and fails to capture intermediate semantic changes. More recently, Reward Forcing\cite{lu2025reward} introduced EMA-Sink, compressing all historical information without distinction, inevitably bluring distinct semantic features, particularly during sudden motion changes or scene transitions. 
\begin{figure*}[!t]
  \centering
  \includegraphics[width=\linewidth]{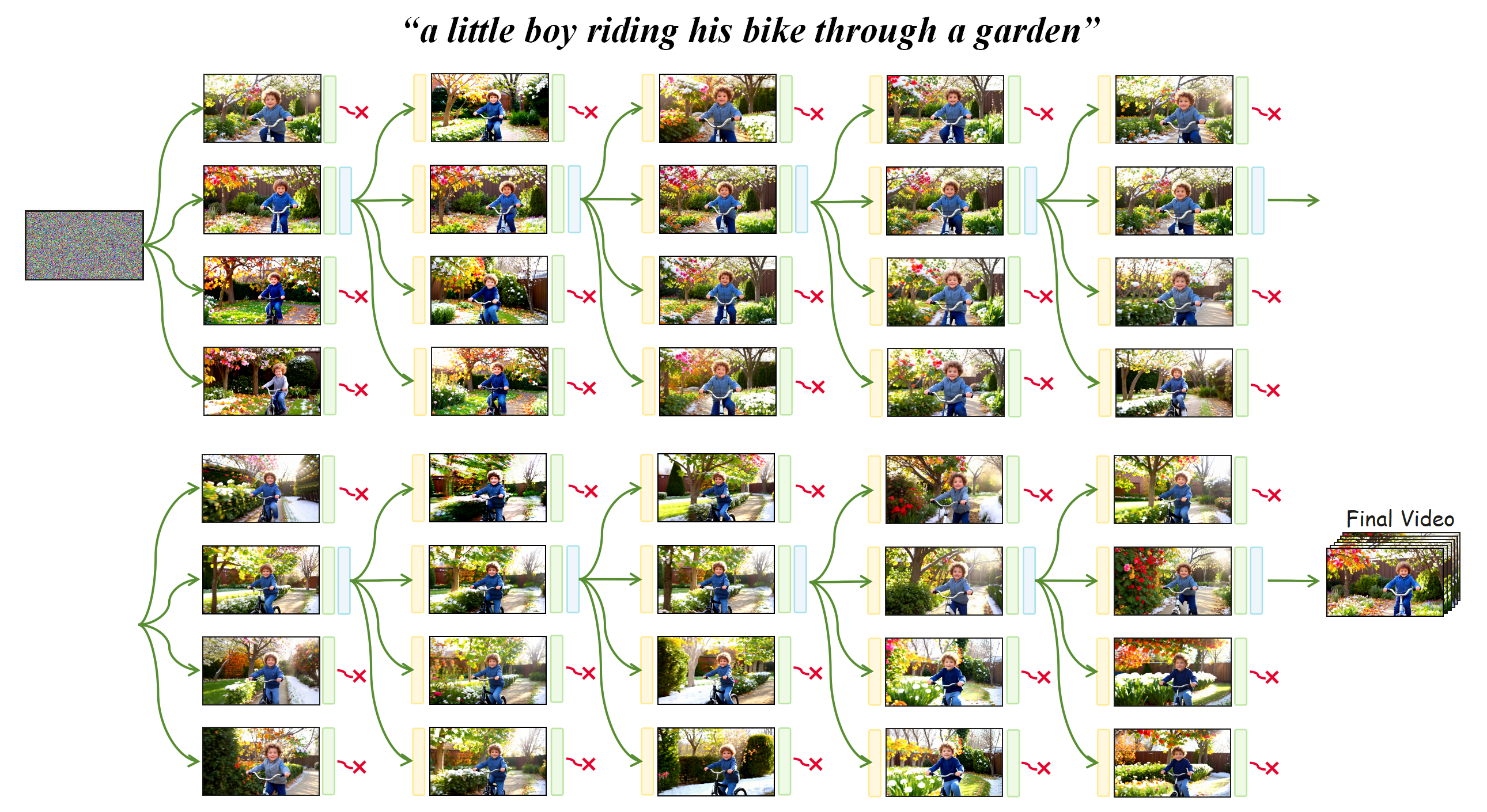}
  \caption{\textbf{The complete search path of the case in the Figure \ref{fig:pipeline}. } }
  \label{fig:search}
\end{figure*}
\subsection{Stream‑Scaled Noise Propagation}
\label{sec:noise}
Existing works\cite{zhou2025golden} have shown that the choice of initial noise profoundly impacts the generation quality of models. EvoSearch\cite{he2025scaling} futher demonstrated that neighboring latent states typically share highly similar generation qualities. Building upon this insight within our autoregressive chunk-wise framework, we introduce a Stream‑Scaled Noise Propagation mechanism. This approach focuses on bridging the connection between the initialization noise of the current chunk and the historical noise latents of previously generated high quality video. By capitalizing on this structural correlation, we successfully guide the ongoing synthesis process, leading to significant improvements in video quality.

Rather than randomly sampling the initialization noise $x_{T}^n$ for the n-th chunk from a standard Gaussian distribution, we construct it based on the optimal noise latent from the preceding chunk. This approach effectively establishes temporal dependency, utilizing the historical Gaussian prior to guide the current generation. Specifically, $x_{T}^n$ is initialized via spherical interpolation:
\begin{align}
    x_T^0 &\sim \mathcal{N}(\mathbf{0}, \mathbf{I}), \\
  x_T^n &= \beta x_T^{n-1} + \sqrt{1-\beta^2}\epsilon,\quad \epsilon \sim \mathcal{N}(\mathbf{0}, \mathbf{I}),
\end{align}
where $\beta \in (-1, 1)$ is an interpolation hyperparameter governing the degree of temporal correlation between adjacent chunks. Crucially, this interpolation guarantees that the marginal distribution of the noise remains strictly invariant, consistently adhering to the standard isotropic Gaussian $\mathcal{N}(\mathbf{0}, \mathbf{I})$.


\subsection{Stream‑Scaled Reward Pruning}
\label{sec:reward}
TTS in video generation relies heavily on search algorithms and reward functions, where the latter serves as the crucial compass for navigating the search space. Our method adopts a Beam Search algorithm guided by our well-designed reward function. For each chunk step, we maintain a beam of $K$ viable candidates. Each of these $K$ candidates is then expanded by generating $M$ alternative next chunks, resulting in a newly expanded pool of $K \times M$ candidates. Based on the evaluative feedback from the reward function, we prune the search space by selecting the top-K candidates to carry forward. 
Tailored to the inherent chunk-by-chunk generation paradigm of streaming video, we propose Stream‑Scaled Reward Pruning. This mechanism preserves the high-fidelity aesthetic quality of local short sequences, while simultaneously enforcing the overarching temporal coherence across global long sequences. 

Specifically, the evaluation of the generated n-th chunk $x_0^n$ is decoupled into two complementary components. The short score is derived by applying an image reward model to independently assess all frames comprising the chunk. Conversely, the long score is computed over an extended temporal context; by incorporating a sliding window, a video reward model evaluates the sequence within the window, comprehensively factoring in text alignment, visual quality, and motion coherence. 
\begin{align}
  S^n_{\text{short}} &= \frac{1}{F}\sum_{f=1}^F\text{ImageReward}(x_0^n[:,f]), \\
  S^n_{\text{long}} &= \text{VideoReward}(\text{output}[: , \text{max}(0, n - w + 1):n]),
\end{align}
where $F$ denotes the total number of frames within chunk $x_0^n$, and $w$ represents the size of the sliding window for long sequence evaluation, $output[: , max(0,n - w + 1):n]$ explicitly denotes the concatenated video sequence comprising the most recently generated chunks within the sliding window. 

Through above formulation, $S^n_{short}$ captures the average spatial fidelity across individual frames, while $S^n_{long}$ holistically assesses the temporal coherence over a broader contextual horizon. To achieve the optimal balance between local aesthetics and global coherence, we design a dynamic weighted fusion strategy with threshold constraint. The weight assigned to the short sequence score linearly increases based on the absolute positional index of the current chunk relative to the entire video length, until it reaches a predefined upper bound where it remains constant. This mechanism dynamically negotiates the trade-off between refining frame-level details and aligning inter-chunk motions. Crucially, the introduction of this threshold constraint avoids frame repetition and stagnation. caused by excessively high short score weight, setting a stable boundary for the balance between spatial fidelity and temporal coherence. 
The final score for the n-th chunk is formulated as: 
\begin{equation}
S^n_{\text{final}} =
\begin{cases}
\frac{n}{N} \cdot S^n_{\text{short}} + (1 - \frac{n}{N} ) \cdot S^n_{\text{long}}, & \frac{n}{N} \le \tau, \\
\tau \cdot S^n_{\text{short}} + (1 - \tau) \cdot S^n_{\text{long}}, & \frac{n}{N} > \tau,
\end{cases}
\end{equation}
where $n$ denotes the index of the current chunk being generated, $N$ represents the total number of chunks in the target video, and $\tau$ is the predefined threshold constraint. 

\subsection{Stream‑Scaled Memory Sinking}
\label{sec:memory}
Fully capitalizing on the inherent strengths of the autoregressive paradigm, we deeply investigate how to effectively harness previously synthesized video context to condition and guide the subsequent  generation. This strategic utilization of historical information is pivotal for ensuring rigorous semantic alignment and temporal coherence throughout the entire video stream. Based on this, We introduce Stream‑Scaled Memory Sinking, a reward-guided dynamic memory updating mechanism that adaptively alternates among discarding, EMA smoothing and appending, ensuring both short term continuity and long term semantic. 
\subsubsection{Semantic Boundary Detection}
To determine the optimal memory updating strategy for the evicted video chunk, we formulate two critical conditions based on the reward score gained from Stream‑Scaled Reward Pruning:

\textbf{Quality Gate:}We first ensure that only high quality chunks are introduced into the global sink. We define the quality condition as: 
\begin{equation}
  \mathcal{C}_{\text{quality}}:=S^{n}_{\text{short}}-\overline{S}_{\text{short}} > \tau _{\text{short}} , 
\end{equation}
where $S^{n}_{\text{short}}$ is the image reward score of the n-th video chunk and $\overline{S}_{\text{short}}$ denotes the moving average of historical short scores, and $\tau _{\text{short}}$ is a predefined threshold. Satisfying this condition guarantees that the KV-cache to be stored possesses sufficient generation quality without visual degradation.

\textbf{Transition Detector:}
To identify scene transitions or significant motion changes, we monitor the fluctuation of the long-term video reward. The transition condition is defined as: 
\begin{equation}
  \mathcal{C}_{\text{transition}}:=S^{n-1}_{\text{long}}-S^{n}_{\text{long}} > \tau _{\text{long}} ,
\end{equation}
where $S^{n}_{\text{long}}$ is the temporal coherence score of the n-th video chunk concatenated with historical chunks, and $S^{n-1}_{\text{long}}$ is the score from the previous chunk. A significant drop exceeding the threshold $\tau _{\text{long}}$ indicates a disruption in temporal continuity, implying the emergence of a new scene or drastic action. 
\subsubsection{Dynamic Memory Mechanism}
Our baseline model, LongLive, employs a sliding window mechanism equipped with attention sinks for context management. Formally, assuming the KV cache has a sliding window size of $w$, when synthesizing the chunk $x_{n+w}$, the key-value pair$(K^{n},V^{n})$ corresponding to the oldest chunk $x_{n}$ is shifted out of the window and naively discarded. Instead of unconditionally discarding the evicted chunk, our method adaptively determines its specific handling operation based on the feedback derived from the Stream‑Scaled Reward Pruning. 

Specifically, we dynamically route $(K^{n},V^{n})$ through one of the following three distinct updating pathways: 
\textbf{Discard}
If the evicted chunk $x_{n}$ fails to pass the quality gate ($\neg \mathcal{C}_{\text{quality}}$), to prevent error accumulation and context pollution, we permanently discard $(K^{n},V^{n})$, regardless of its temporal coherence. 
\begin{align}
  S^{n+w}_K &= S^{n+w-1}_K, \\
  S^{n+w}_V &= S^{n+w-1}_V.
\end{align}
\textbf{EMA-Sink}
When the evicted chunk $x_{n}$ possesses high spatial quality but triggers no semantic transition ($\mathcal{C}_{\text{quality}} \land \neg \mathcal{C}_{\text{transition}}$), it indicates smooth, continuous motion within the same scene, implying that the 
n-th chunk contains excessive redundant information overlapping with the preceding video content. To preserve this local temporal coherence without incurring additional GPU memory overhead, we integrate $(K^{n},V^{n})$ into the latest sink: 
\begin{align}
  S^{n+w}_K &= [S^{n+w-1}_K[:-1]; \alpha \cdot S^{n+w-1}_K[-1:] + (1-\alpha) \cdot K^{n}], \\
  S^{n+w}_V &= [S^{n+w-1}_V[:-1]; \alpha \cdot S^{n+w-1}_V[-1:] + (1-\alpha) \cdot V^{n}],
\end{align}
where $\alpha \in (0, 1)$ is a fixed decay factor. This Sink-EMA operation acts as a continuous update mechanism for the Short Memory, compressing redundant sequential chunks into a stable, fixed-size representation.

\textbf{Append-Sink}
If the high-quality chunk coincides with a significant drop in long term coherence ($\mathcal{C}_{\text{quality}} \land \mathcal{C}_{\text{transition}}$), it serves as a strong indicator of a semantic shift, such as a scene shift or a drastically new action. Applying naive EMA in this scenario would indiscriminately blend distinct visual features, causing semantic blurring and feature corruption. Therefore, we directly append $(K^{n},V^{n})$ to the global sink sequence as a new, discrete anchor: 
\begin{align}
  S^{n+w}_K &= [S^{n+w-1}_K;K^{n}], \\
  S^{n+w}_V &= [S^{n+w-1}_V;V^{n}].
\end{align}
During attention computation, we concat the latest sink states with the local window context:
\begin{align}
  K^{n+w}_{global} &= [S^{n+w}_K;K^{n+1:n+w}], \\
  V^{n+w}_{global} &= [S^{n+w}_V;V^{n+1:n+w}].
\end{align}
By dynamically alternating among discarding, EMA and appending guided by reward feedback, our mechanism effectively decouples short term continuity from long term memory preservation. It overcomes both the rigid frame copying issue of static sinks and the feature corruption caused by uniform EMA blending, paving the way for consistent, long video generation. 

\section{Experiments}
\label{sec:Experiments}

\subsection{Experiments Settings}
\subsubsection{Implementation Details}
We evaluate our scaling method on LongLive\cite{yang2025longlive}, which is built on Wan2.1-T2V-1.3B\cite{wang2025wan}. Following LongLive, the initial KV cache strategy is set an attention window size of 9 alongside a sink size of 3. The feedback mechanism in our pipeline is instantiated as a dual-level evaluation framework. Specifically, for short-sequence evaluation, we adopt established image reward models (e.g., HPSv3\cite{ma2025hpsv3}, ImageReward\cite{xu2023imagereward}, MHP\cite{zhang2024learning}) to assess frame-level spatial aesthetics and visual fidelity. Conversely, for long-sequence evaluation, we employ state-of-the-art video reward models (e.g., VisionReward\cite{xu2026visionreward}, VideoAlign\cite{liu2025improving}, VideoLLaMA3\cite{zhang2025videollama}). These video-based models are applied over a sliding window of 10 chunks to holistically measure temporal coherence and global dynamics within extended contexts. We use random seed 42 for all experiments, and each generated video at 16 FPS with a resolution of 832 × 480. 
\subsubsection{Evaluation Benchmark}
To comprehensively validate the effectiveness of our method, we conduct systematic evaluations on both short (5s) and long (30s) video generation. The 5s videos are synthesized using a comprehensive set of 946 prompts from VBench\cite{huang2024vbench,huang2025vbench++}, whereas the 30s videos are generated using the first 128 prompts from MovieGen\cite{polyak2024movie}. We evaluate the 5s and 30s videos using VBench\cite{huang2024vbench} and VBench-long\cite{huang2025vbench++} respectively, measuring Subject Consistency, Background Consistency, Motion Smoothness, Aesthetic Quality, and Imaging Quality. Additionally, to capture human visual preferences, we employ VideoAlign\cite{liu2025improving}, a video reward model built upon Qwen2-VL-2B\cite{wang2024qwen2} and trained on a dataset of multi-faceted human preferences to align automated video quality evaluation with human perception, to evaluate the videos across three dimensions: visual quality(VQ), motion quality(MQ), and text alignment(TA).
\subsection{Comparison with state-of-the-art methods}
\subsubsection{Quantitative Results}
To comprehensively evaluate the effectiveness of our proposed Stream-T1, we compare against three representative open-source models: CausVid\cite{yin2025slow}, Self-forcing\cite{huang2025self}, LongLive\cite{yang2025longlive} on both 5s and 30s video generation. Overall, extensive experimental analyses demonstrate that Stream-T1 significantly and comprehensively elevates the temporal consistency, motion coherence, and frame-level visual fidelity of the generated videos.

On 5s clips shown in Tab. \ref{tab:5s}, Stream-T1 records the best results on six quality metrics(Subject Consistency, Background Consistency, Motion Smoothness, Aesthetic Quality, MQ and TA) and ranks the second on Imaging Quality and VQ. 
\begin{table*}[h]
  \centering
  \caption{Quantitative comparison with baseline of 5s video generation. The best results are highlighted in bold}
  \label{tab:5s}
  \resizebox{\linewidth}{!}{
  \begin{tabular}{ccccccccc}
    \toprule
    \multirow{2}{*}{\textbf{Method}} & 
    \multicolumn{5}{c}{\textbf{VBench$\uparrow$}} & 
    \multicolumn{3}{c}{\textbf{VideoAlign$\uparrow$}} \\
    \cmidrule(lr){2-6} \cmidrule(lr){7-9}
    & \textbf{\makecell{Subject\\Consistency}} & \textbf{\makecell{Background\\Consistency}} &  \textbf{\makecell{Motion\\Smoothness}} & \textbf{\makecell{Imaging\\Quality}} & \textbf{\makecell{Aesthetic\\Quality}} & \textbf{VQ} & \textbf{MQ} & \textbf{TA}\\
    \midrule
    \small{CausVid\cite{yin2025slow}} & 96.33 & 95.56 & 98.66 & 69.69 & 62.90 & \textbf{0.433} & 0.550 & 1.02 \\
    \small{Self-forcing\cite{huang2025self}} & 95.26 & 95.67 & 98.67 & \textbf{71.61} & 63.97 & 0.099 & 0.088 & 1.193 \\
    \small{LongLive\cite{yang2025longlive}}& 97.00 & 96.78 & 99.12 & 71.28 & 65.28 & 0.285 & 0.350 & 1.193 \\
    \textbf{Stream-T1 (on LongLive)} & \textbf{97.25} & \textbf{97.05} & \textbf{99.15} & 71.42 & \textbf{65.98} & 0.426 & \textbf{0.629} & \textbf{1.305} \\
    & $\triangle$0.26\% & $\triangle$0.28\% & $\triangle$0.03\% & $\triangle$0.2\% & $\triangle$1.07\% & $\triangle$49.47\% & $\triangle$79.71\% & $\triangle$9.39\% \\
    \bottomrule
\end{tabular}
}
\end{table*}

On 30s videos shown in Tab. \ref{tab:30s}, Stream-T1 outperforms state-of-the-art baselines across almost all metrics. Stream-T1 achieves the highest scores in Subject Consistency, Background Consistency, Motion Smoothness, Imaging Quality, and Aesthetic Quality. This superiority is further corroborated by the human-aligned VideoAlign, where our model obtains the best VQ and TA. And Stream-T1 gets second score in MQ.

\begin{table*}[h]
  \centering
  \caption{Quantitative comparison with baseline of 30s video generation. The best results are highlighted in bold}
  \label{tab:30s}
  \resizebox{\linewidth}{!}{
  \begin{tabular}{ccccccccl}
    \toprule
    \multirow{2}{*}{\textbf{Method}} & 
    \multicolumn{5}{c}{\textbf{VBench Long$\uparrow$}} & 
    \multicolumn{3}{c}{\textbf{VideoAlign$\uparrow$}} \\
    \cmidrule(lr){2-6} \cmidrule(lr){7-9}
    & \textbf{\makecell{Subject\\Consistency}} & \textbf{\makecell{Background\\Consistency}} &  \textbf{\makecell{Motion\\Smoothness}} & \textbf{\makecell{Imaging\\Quality}} & \textbf{\makecell{Aesthetic\\Quality}} & \textbf{VQ} & \textbf{MQ} & \textbf{TA}\\
    \midrule
    \small{CausVid} & 97.91 & 96.74 & 98.15 & 66.32 & 59.71 & -0.144 & \textbf{0.328} & 0.501 \\
    \small{Self-forcing} & 97.18 & 96.37 & 98.35 & 68.35 & 59.19 & -0.461 & -0.216 & 0.656 \\
    \small{LongLive}& 97.90 & 96.82 & 98.78 & 68.99 & 61.56 & -0.169 & -0.002 & 1.073\\
    \textbf{Stream-T1 (on LongLive)} & \textbf{98.43} & \textbf{97.18} & \textbf{99.03} & \textbf{69.10} & \textbf{62.11} & \textbf{-0.073} & 0.226 & \textbf{1.170}\\
     & $\triangle$0.54\% & $\triangle$0.37\% & $\triangle$0.25\% & $\triangle$0.16\% & $\triangle$0.89\% & $\triangle$56.8\% & $\triangle$11400\% & $\triangle$9\% \\
    \bottomrule
\end{tabular}
}
\end{table*}
\subsubsection{Qualitative Results}
\begin{figure*}[!htb]
  \centering
  \includegraphics[width=\linewidth]{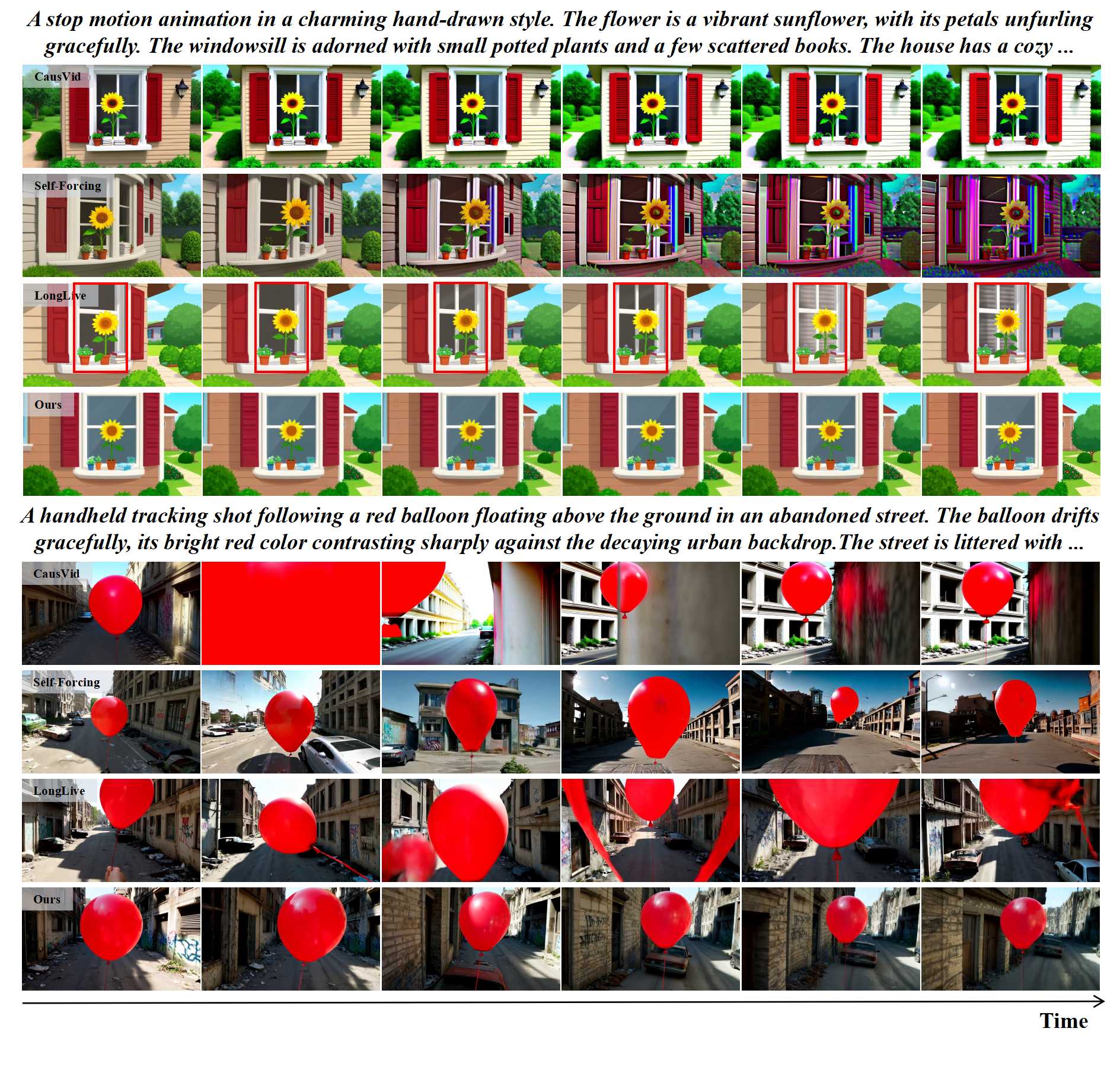}
  \caption{\textbf{Qualitative comparisons of Stream-T1 with Causvid, Self-Forcing, and LongLive. }Stream-T1 significantly elevates the temporal consistency and frame-level visual fidelity of the generated videos.}
  \label{fig:results}
\end{figure*}

As shown in Fig. \ref{fig:results}, with increasing video length, existing baseline models suffer from severe quality degradation. Specifically, CausVid\cite{yin2025slow} and Self-Forcing\cite{huang2025self} encounter severe frame-level visual distortion in long-sequence generation. Although LongLive\cite{yang2025longlive} mitigates spatial distortion to some extent, it experiences a drastic drop in temporal consistency. In contrast, Stream-T1 demonstrates remarkable long-term stability, consistently maintaining high spatiotemporal coherence and superior visual aesthetics throughout the entire 30s video.
\subsection{Comparison with Test-Time Scaling}
We contrast our approach with standard test-time scaling algorithms, specifically Best-of-N and Beam Search. While these methods attempt to mitigate generation errors through redundant candidate sampling or iterative exploration, they fundamentally operate under a passive selection paradigm, merely picking the most sub-optimal outcome from a fixed generated pool. In contrast, our framework shifts to an active optimization paradigm. Rather than merely expanding the search space, we actively guide the video generation process by dynamically refining both the latent noise and the context memory. As demonstrated by the quantitative results in Tab. \ref{tab:tts}, this active intervention strategy exhibits profound superiority, achieving state-of-the-art performance across all evaluative metrics.
\begin{table*}[h]
  \centering
  \caption{Quantitative comparison with test-time scaling methods of 30s video generation. Stream-T1 achieving state-of-the-art performance across all evaluative metrics.}
  \label{tab:tts}
  \resizebox{\linewidth}{!}{
  \begin{tabular}{cccccccccl}
    \toprule
    \multirow{2}{*}{\textbf{Method}} & 
    \multicolumn{5}{c}{\textbf{VBench Long$\uparrow$}} & 
    \multicolumn{3}{c}{\textbf{VideoAlign$\uparrow$}} \\
    \cmidrule(lr){2-6} \cmidrule(lr){7-9}
    & \textbf{\makecell{Subject\\Consistency}} & \textbf{\makecell{Background\\Consistency}} & \textbf{\makecell{Motion\\Smoothness}} & \textbf{\makecell{Imaging\\Quality}} & \textbf{\makecell{Aesthetic\\Quality}} & \textbf{VQ} & \textbf{MQ} & \textbf{TA}\\
    \midrule
    \small{LongLive}& 97.90 & 96.82 & 98.78 & 68.99 & 61.56 & -0.169 & -0.002 & 1.073\\
    \small{$+$ \small{Best of N}}& 98.13 & 96.88 & 98.86 & 69.34 & 61.97 & -0.083 & 0.062 & 1.160\\
    \small{$+$ \small{BeamSearch}}& 98.28 & 97.03 & 98.90 & 69.05 & 61.85 & -0.077 & 0.165 & 1.159\\
    $+$ \textbf{Ours} & \textbf{98.43} & \textbf{97.18} & \textbf{99.03} & \textbf{69.10} & \textbf{62.11} &\textbf{ -0.073} & \textbf{0.226} & \textbf{1.170}\\
    \bottomrule
\end{tabular}
}
\end{table*}
\subsection{Ablations}
To thoroughly validate the individual contributions of our proposed components, we conduct comprehensive ablation studies on the 30s video generation using the first 128 prompts from MovieGen. Both quantitative metrics and extensive qualitative visual comparisons consistently demonstrate the necessity and efficacy of our designed modules.
\subsubsection{Qualitative experiment on each component}
\begin{figure*}[htbp]
  \centering
  \includegraphics[width=\linewidth]{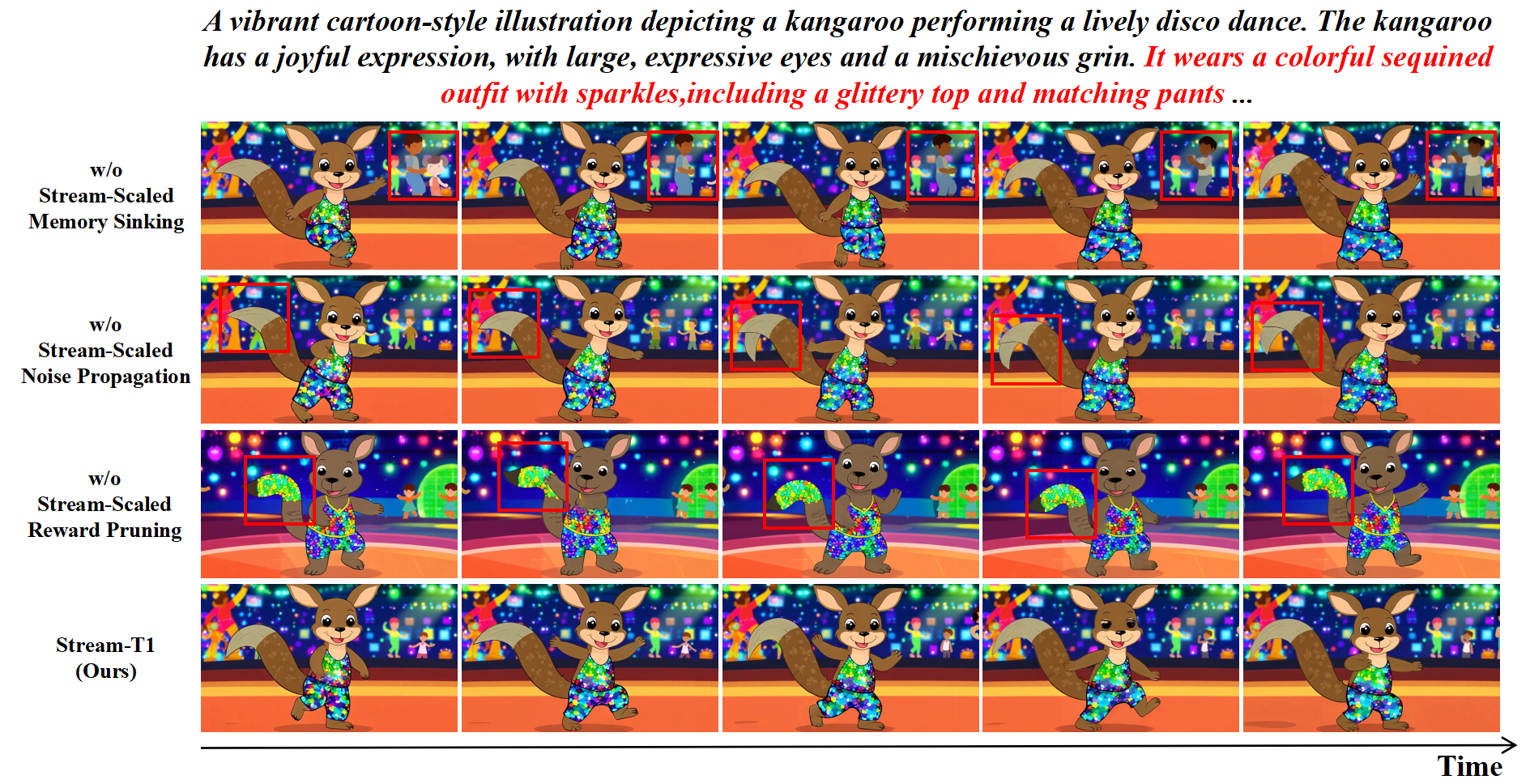}
  \caption{\textbf{Qualitative ablations studies on each component. }Omitting the Stream-Scaled Memory Sinking degrades background stability. Removing the Stream-Scaled Noise Propagation introduces local structural artifacts (e.g., on the subject's tail). Eliminating the Stream-Scaled Reward Pruning leads to distinct semantic misalignment and deteriorated aesthetic quality.}
  \label{fig:ablation}
\end{figure*}

As illustrated by the qualitative results in Figure \ref{fig:ablation}, omitting the Stream-Scaled Memory Sinking degrades background stability. Removing the Stream-Scaled Noise Propagation introduces local structural artifacts (e.g., on the subject's tail). Finally, eliminating the Stream-Scaled Reward Pruning leads to distinct semantic misalignment and deteriorated aesthetic quality. These findings confirm that each component is essential for high-quality video generation.
\subsubsection{Quantitative experiment on each component}
The ablation study in Tab. \ref{tab:abalation} shows that removing the Stream‑Scaled Memory Sinking leads to a noticeable gain in Imaging Quality, it severely compromises Subject and Background Consistency. This sharp deterioration effectively validates the crucial role of our dynamic context management in sustaining robust long-term temporal coherence over extended video sequences. Furthermore, ablating the Stream‑Scaled Noise Propagation results in a uniform performance drop across all metrics, validating its necessity for overall stability. Lastly, eliminating the Stream‑Scaled Reward Pruning marginally increases Imaging Quality but causes a drastic decline in all other metrics, explicitly emphasizing the necessity of our long-short combined reward in balancing spatial and temporal dynamics.
\begin{table*}[h]
  \centering
  \caption{Quantitative ablation studies on each component of 30s video generation. }
  \label{tab:abalation}
  \resizebox{\linewidth}{!}{
  \begin{tabular}{ccccccccl}
    \toprule
    \multirow{2}{*}{\textbf{Method}} & 
    \multicolumn{5}{c}{\textbf{VBench Long$\uparrow$}} & 
    \multicolumn{3}{c}{\textbf{VideoAlign$\uparrow$}} \\
    \cmidrule(lr){2-6} \cmidrule(lr){7-9}
    & \textbf{\makecell{Subject\\Consistency}} & \textbf{\makecell{Background\\Consistency}} &  \textbf{\makecell{Motion\\Smoothness}} & \textbf{\makecell{Imaging\\Quality}} & \textbf{\makecell{Aesthetic\\Quality}} & \textbf{VQ} & \textbf{MQ} & \textbf{TA}\\
    \midrule
     \small{w/o Stream‑Scaled Memory Sinking}& 98.30 & 97.04 & 98.92 & \textbf{69.51} & 61.90 & -0.083 & 0.188 & 1.146\\
      \small{w/o Stream‑Scaled Noise Propagation}& 98.35 & 97.14 & 98.98 & 69.07 & 61.99 & -0.094 & 0.176 & 1.164\\
    \small{w/o Stream‑Scaled Reward Pruning}& 98.04 & 96.88 & 98.87 & 69.17 & 61.22 & -0.173 & 0.014 & 1.035\\
    \textbf{Ours} & \textbf{98.43} & \textbf{97.18} & \textbf{99.03} & 69.10 & \textbf{62.11} & \textbf{-0.073} & \textbf{0.226} & \textbf{1.170}\\
    \bottomrule
\end{tabular}
}
\end{table*}

\section{Conclusion}
\label{sec:Conclusion}
In this paper, we introduced \textbf{Stream-T1}, a novel TTS framework tailored for streaming video generation. Stream-T1 operates through three distinct phases. First, Stream-Scaled Noise Propagation actively refines the initial latent noise of the generating chunk using historically proven, high-quality previous chunk noise. Upon generation, Stream-Scaled Reward Pruning module conducts a comprehensive evaluation fusing short-term spatial and long-term temporal metrics. This holistic feedback is also to derive dynamic memory updating signals. These signals govern the KV-cache sink updates, guiding the context management for subsequent video generation. Evaluated on comprehensive benchmarks, Stream-T1 significantly improves temporal consistency, motion smoothness, and frame-level visual quality.

\clearpage

\bibliographystyle{plainnat}
\bibliography{main}

\end{document}